# Measuring Behavioural Similarity of Cellular Automata

Peter D. Turney[*]

## Abstract

Conway's Game of Life is the best-known cellular automaton. It is a classic model of emergence and self-organization, it is Turing-complete, and it can simulate a universal constructor. The Game of Life belongs to the set of semi-totalistic cellular automata, a family with 262,144 members. Many of these automata may deserve as much attention as the Game of Life, if not more. The challenge we address here is to provide a structure for organizing this large family, to make it easier to find interesting automata, and to understand the relations between automata. Packard and Wolfram (1985) divided the family into four classes, based on the observed behaviours of the rules. Eppstein (2010) proposed an alternative four-class system, based on the forms of the rules. Instead of a class-based organization, we propose a continuous high-dimensional vector space, where each automaton is represented by a point in the space. The distance between two automata in this space corresponds to the differences in their behavioural characteristics. Nearest neighbours in the space have similar behaviours. This space should make it easier for researchers to see the structure of the family of semi-totalistic rules and to find the hidden gems in the family.

**Keywords:** cellular automata, semi-totalistic automata, outer-totalistic automata, Conway's Game of Life, similarity of cellular automata, behaviour of cellular automata.

Letter (about 2,000 words)

[*] Ronin Institute, 127 Haddon Place, Montclair, NJ 07043-2314, USA,
peter.turney@ronininstitute.com, 819-661-4625.

# 1 Introduction

The *Game of Life* (GoL) is a solitaire game invented by John Conway and introduced to the world by Martin Gardner in *Scientific American* [3]. It is played on a potentially infinite, two-dimensional grid of square cells. Each cell is either *dead* (state 0) or *alive* (state 1). The state of a cell changes with time, based on the states of its eight nearest neighbours (called the *Moore neighbourhood*). Time passes in a series of discrete intervals. At time $t = 0$, the player of the game chooses the initial states of the grid. The initial states form a *seed pattern* that determines the course of the game. The states at time $t$ uniquely determine the states at time $t + 1$. With each increment of $t$, all of the cells are updated. As the game runs, patterns grow and decay, resembling living organisms.

The rule for changing states in GoL can be compactly represented as B3/S23, where B means *born* and S means *survives*. A cell is *born* (it switches from state 0 to state 1) when exactly three of its eight nearest neighbours are alive (in state 1). A cell *survives* (remains in state 1) when it has either two or three living neighbours. Otherwise, a cell *dies* (it switches to state 0 or remains in state 0).

GoL is a *cellular automaton*, a discrete, abstract computational system. It is the best-known member of the family of cellular automata. It is popular as a model of emergence and self-organization [1], it is Turing-complete [8], and it is a universal constructor [5].

GoL is a member of the family of *semi-totalistic cellular automata* (also called *outer-totalistic*). The rules for this family have the general form B$x$/S$y$, where $x$ and $y$ are generated by deleting digits from the string 012345678, including deleting no digits or deleting all digits [2]. Since there are nine digits available for B and nine digits available for S, there are 2 to the power of 18 (262,144) possible semi-totalistic rules.

Packard and Wolfram [7] introduced a four-class system for characterizing two-dimensional cellular automata. Eppstein [2] an alternative four-class system, based on analysis of the B$x$/S$y$ rule forms. These classification systems are interesting and useful, but a researcher in the field of cellular automata might hope for more guidance than four classes can provide. One quarter of 262,144 is 65,536, which leaves a large space to explore.

In this letter, we introduce a continuous high-dimensional vector space, where the statistical behaviour of an automaton corresponds to a point in the space. For a given semi-totalistic rule, such as B3/S23, we create an initial random soup of living cells and then run the game. We then randomly sample cells in the game to estimate the probabilities of the state transitions, which depend on both the rules of the game and the particular patterns of neighbours that tend to arise in the game. The behaviour is then represented as a vector of the estimated probabilities of each possible state transition.

With the probability vectors for all 262,144 semi-totalistic cellular automata, we can measure the behavioural similarity of any two automata by any suitable distance measure. In this letter, we have chosen to use Euclidean distance, where low distance corresponds to high behavioural similarity. This continuous high-dimensional vector space enables many ways of searching for interesting automata and finding relations among automata:

**Find more like this:** Given an automaton that is interesting, find similar automata that might also be interesting, by sorting the automata in order of increasing distance from the given automaton.

**Finding hybrids:** Given two different automata, find a third automaton that is half way between them.

**Finding models for phenomena:** Given some natural phenomenon that looks like it might be something a cellular automaton could model, try to make a behavioural vector for it, then use the behavioural vector to find the best automaton for modeling the phenomenon.

**Unsupervised clustering:** Use a similarity-based clustering algorithm to cluster automata by vector similarity.

**Supervised clustering:** Use vector similarity for supervised clustering, given some manually identified examples of each desired cluster.

**Finding opposites:** Given an automaton, find the automaton that is least similar to the given rule; the maximally distant automaton.

**Finding idiosyncrasy:** For each automaton, find its nearest neighbour, then make note of how different the two automata are; output the automaton that is least like its nearest neighbour (most unique).

**Finding paradigmatic examples:** Given a group of automata that belong to the same cluster, find the automaton that is the centroid or average of the cluster, to serve as a paradigmatic example of the cluster.

**Projection into subspaces:** To understand the high-dimensional vector space better, project it into lower dimensional subspaces, such as two-dimensional subspaces, which are easier to visualize.

The source code for calculating the high-dimensional vectors is available for downloading, along with the vectors for all 262,144 possible semi-totalistic rules [10]. The source code uses Python and Golly [9].

In Section 2, we describe the problems of strobing and infinity that arise with rules that contain B0. In Section 3, we present a 36-dimensional space for representing the behaviour of automata, ignoring the issues of strobing and infinities, in order to simplify the presentation. In Section 4, we address the problems of strobing and infinity, using a 72-dimensional space. We conclude in Section 5.

## 2 The Problems of Strobing and Infinity

In a typical game, we begin at time $t = 0$ with a potentially infinite grid and a finite number of live cells. If the given rule contains B0, all dead cells with no neighbours will come alive at time $t = 1$. This means there will be an infinite number of live cells at $t = 1$. If the rule does not contain S8, at time $t = 2$, an infinite number of cells will die. This creates an annoying strobing effect, with alternating light and dark images at odd and even times.

The popular Golly software for cellular automata automatically corrects the problems with strobing and infinities [4]. If a rule contains B0 and S8, then the rule is replaced with an equivalent rule that avoids the problem of an infinite number of live cells. If a rule contains B0 but not S8, then the rule is replaced with two rules, one for even times and one for odd times, to avoid the problem of strobing. We do not have the space here to explain how Golly adjusts the rules, but this information is available on the Golly website [4] and our Python code includes the required adjustments [10].

## 3 The 36-dimensional Space for Semi-totalistic Automata

As an example of the 36-dimensional space for semi-totalistic automata, Table 1 shows two vectors for the Game of Life, B3/S23. The first is a Boolean vector that expresses the rule B3/S23 with four 9-dimensional vectors, Born (B), Survive (S), Unborn (U), and Die (D). The rule tells us which state transitions are *allowed* (1) and which transitions are *forbidden* (0). We can see that the Boolean vector for U is the inverse of the vector for B, and D is the inverse of S. A limitation of measuring distance with Boolean vectors is that any two rules that differ by the insertion or deletion of $N$ numbers are exactly the same distance apart. The Boolean vector space is not capable of making fine distinctions between rules.

> Insert Table 1 here. Tables are at
> the end of this document.

The second vector is a real-valued vector of probabilities, such that the sum of the probabilities is 1. This vector tells us the estimated probability of each *allowed* state transition, based on many observations of the Game of Life. State transitions that are *forbidden* have a probability of zero. Table 1 shows that a cell with two neighbours is more likely to survive than a cell with three neighbours (probability 0.0414 for two and probability 0.0327 for three). This implies that a line of live cells is slightly more likely to continue than to branch out. The probability vectors give us qualitative behavioural information that is not available in the Boolean vectors.

The probabilities in Table 1 were generated by repeatedly making random soups and running them to see how they evolve. Table 2 shows the parameter settings we used for these experiments. Taking a sample consists of randomly selecting a cell (alive or dead) and inspecting its state and the state of its neighbours at time $t$ and then inspecting its new state at time $t + 1$, to see what transition took place.

> Insert Table 2 here. Tables are at the end of this document.

The initial random soup is contained in a 16×16 square of cells (initial_size), following the example of the most popular software for exploring random soups, Catagolue [5]. The soup is generated in two steps. First, we randomly select a number $d$ using a continuous uniform distribution between 0 and 1 (density_range). This $d$ gives us the desired density of live cells for the 16×16 square. Second, we iterate through the cells in the square, randomly assigning state 1 with probability $d$ and state 0 with probability $1-d$. We then run the soup for 50 steps, to see how it develops (num_steps). We then take 50 samples of the state transitions (num_samples). This process is repeated 1000 times (num_trials), each time with a different density $d$. From the 50,000 state transitions (1000 soups with 50 samples per soup), we estimate the state transition probabilities for the real-valued probability vectors.

After 50 steps (num_steps), the random soup may not have settled into a stable configuration, but we believe that some degree of instability is natural, so it is not necessary to run the game until it is completely stable. It is also advantageous to keep the number of steps relatively small, due to the time required for computation, since we are dealing with 262,144,000 soups (num_trials × $2^{18}$).

## 4 The 72-dimensional Space for Semi-totalistic Automata

As an example of the 72-dimensional space for semi-totalistic automata, Table 3 shows the vectors for the rule B03/S23. This rule causes strobing, because it contains B0 and not S8 (see Section 2), so we replace it with two 36-dimensional Boolean vectors, one for even values of $t$ (B1245678/S0145678) and one for odd values of $t$ (B56/S58) [4]. To express these anti-strobing rules in a vector space, we need to join these two 36-dimensional Boolean vectors, creating a 72-dimensional Boolean vector. The bottom eight rows of the table show the corresponding 72-dimensional real-valued vector of probabilities. The probabilities in the first 36 dimensions sum to 1 and the probabilities in the second 36 dimensions also sum to 1, so the whole vector sums to 2.

> Insert Table 3 here. Tables are at the end of this document.

Rules without strobing do not require 72 dimensions, but we can easily expand them to 72-dimensions by repeating the 36-dimensional vector. The 72-dimensional vector has an even part and an odd part, so it is natural to duplicate the 36-dimensional vector when the same rule is used for even and odd times. This allows us to compare non-strobing rules (such as B3/S23) with strobing rules (such as B03/S23) in the same 72-dimensional space.

Table 4 shows the nearest neighbours of the rule B3/S23 in the 72-dimensional probability space. There are two cases where Golly modifies rules, in order to prevent strobing and infinity [4]: (1) A rule containing B0 and S8 is black/white reversed to avoid infinity. (2) A rule containing B0 but not S8 is split into two new rules to avoid strobing. In Table 4, case (1) applies: Some of the rules are modified to prevent infinity. This modification gives rise to duplicate rules, which gives us an opportunity to see how much noise there is in our probability estimates. If the probability estimates in the real-valued vectors were perfectly accurate, then the duplicate rules would be adjacent to each other in the table. The duplicates are close together at the top of the list, but they diverge as we go down the list. We will discuss this divergence after we consider another example.

> Insert Table 4 here. Tables are at the end of this document.

Table 5 shows the nearest neighbours of the rule B03/S23 in the 72-dimensional probability space. In Table 5, case (2) applies: All of the rules are modified to prevent strobing. Unlike Table 4, there are no duplicate rules in Table 5.

> Insert Table 5 here. Tables are at the end of this document.

In Table 4 and Table 5, the two final columns show the distances from the target rule in real space and Boolean space. The ranking in column 1 is based on real space alone; the Boolean space is only included for comparison. There are many ties in the Boolean column, which shows that the real space is able to make finer distinctions than the Boolean space. We also see that the Boolean distances yield different rankings from the real-valued distances.

Figure 1 shows the relation between rank and distance for the 200 most highly ranked rules, with one curve for the target rule B3/S23 and a second curve for the target rule B03/S23. The curve for B3/S23 flattens out at about rank 15, and we can see this flattening also in Table 4, in the column of real distances. This flattening explains why the duplicate rules in Table 4 are near each other until we pass rank 15, when the duplicate rules become distant from each other. When the curve is flat, a small amount of noise in the estimated probability can cause a large amount of variation in the estimated rank. However, the curve for B03/S23 does not flatten out as quickly as the curve for B3/S23. Figure 1 suggests that the rankings for B03/S23 will be reliable at least up to rank 40.

> Insert Figure 1 here. The figure is at the end of this document.

# 5 Conclusion

This letter offers a new method for examining the behaviours of the family of semi-totalistic cellular automata and for investigating the relations among the members of the family. Given that there are 262,144 semi-totalistic rules, researchers in the field of cellular automata have a large space to explore. In the spirit of Packard and Wolfram [7] and Eppstein [2], we hope that our method for measuring behavioural similarity of cellular automata will give researchers useful new ways of viewing, organizing, and understanding the semi-totalistic family.

## Acknowledgments

Thanks to the reviewers of *Artificial Life* for their very helpful advice.

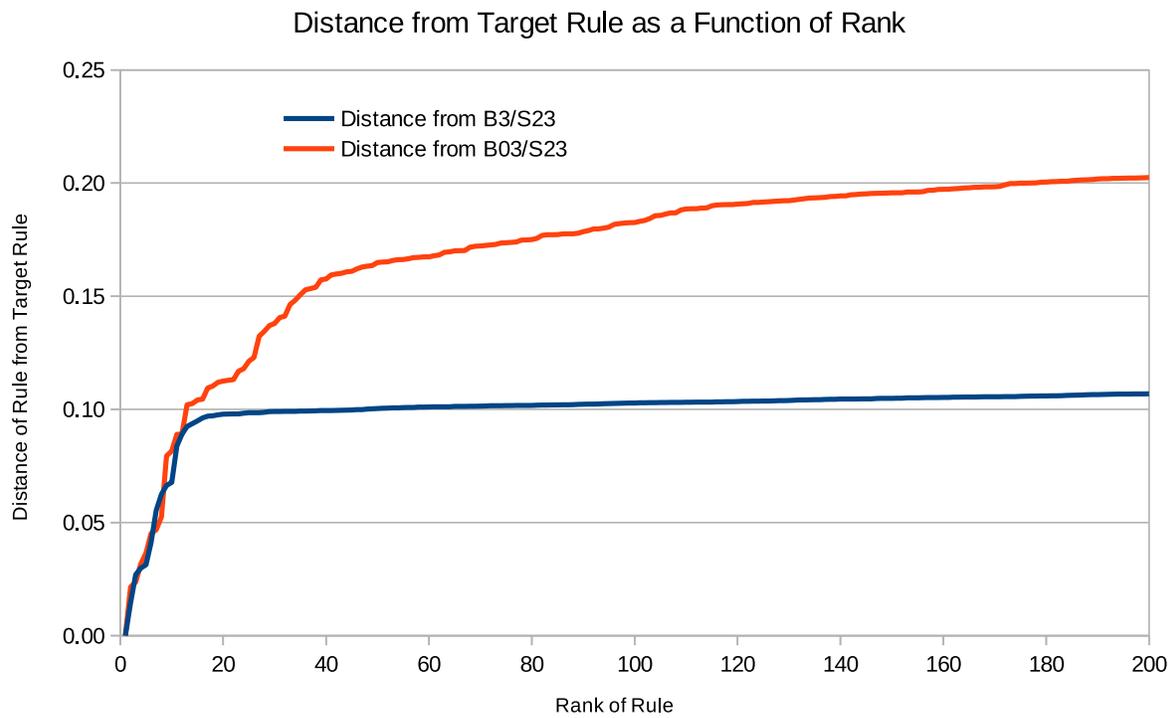

Figure 1. This graph shows how the distance of a given rule from a target rule varies as a function of the rank of the given rule.

Table 1. Here we see two 36-dimensional vectors for the Game of Life, B3/S23. The Boolean vector expresses exactly the same information as the rule, B3/S23, but more explicitly. The real-valued probability vector gives a more detailed picture, showing how the rule behaves in practice.

| Vectors | | Transition | Number of Neighbours | | | | | | | | |
| --- | --- | --- | --- | --- | --- | --- | --- | --- | --- | --- | --- |
| | | $t$ to $t+1$ | 0 | 1 | 2 | 3 | 4 | 5 | 6 | 7 | 8 |
| Boolean | B | 0 to 1 | 0 | 0 | 0 | 1 | 0 | 0 | 0 | 0 | 0 |
| B3/S23 | S | 1 to 1 | 0 | 0 | 1 | 1 | 0 | 0 | 0 | 0 | 0 |
| | U | 0 to 0 | 1 | 1 | 1 | 0 | 1 | 1 | 1 | 1 | 1 |
| | D | 1 to 0 | 1 | 1 | 0 | 0 | 1 | 1 | 1 | 1 | 1 |
| Real | B | 0 to 1 | 0 | 0 | 0 | 0.0454 | 0 | 0 | 0 | 0 | 0 |
| B3/S23 | S | 1 to 1 | 0 | 0 | 0.0414 | 0.0327 | 0 | 0 | 0 | 0 | 0 |
| | U | 0 to 0 | 0.5883 | 0.0963 | 0.1082 | 0 | 0.0212 | 0.0199 | 0.0037 | 0.0003 | 0.0002 |
| | D | 1 to 0 | 0.0013 | 0.0149 | 0 | 0 | 0.0160 | 0.0073 | 0.0025 | 0.0004 | 0.0001 |

Table 2. This table shows the parameter settings we used for the experiments that estimated the probabilities in the real-valued vectors.

| Parameter | Value | Description |
| --- | --- | --- |
| density_range | [0.0, 1.0] | density range for initial random soup matrix |
| initial_size | 16 | width and height for initial random soup matrix |
| num_steps | 50 | number of steps for a run of the cellular automaton |
| num_samples | 50 | number of samples to collect from each run |
| num_trials | 1000 | number of trials (runs) to evaluate for each rule |

Table 3. Here we have vectors for the strobing rule B03/S23. First, we have the 36-dimensional Boolean vector for B03/S23. Second, we have two 36-dimensional Boolean vectors, one for even $t$ and one for odd $t$, which combine to create a 72-dimensional Boolean vector that does not strobe. Third, we have two 36-dimensional probability vectors, which combine to create a 72-dimensional probability vector.

| Vectors | Transition | | | | Number of Neighbours | | | | | |
|---|---|---|---|---|---|---|---|---|---|---|
| | $t$ to $t+1$ | 0 | 1 | 2 | 3 | 4 | 5 | 6 | 7 | 8 |
| Boolean | B   0 to 1 | 1 | 0 | 0 | 1 | 0 | 0 | 0 | 0 | 0 |
| B03/S23 | S   1 to 1 | 0 | 0 | 1 | 1 | 0 | 0 | 0 | 0 | 0 |
| | U   0 to 0 | 0 | 1 | 1 | 0 | 1 | 1 | 1 | 1 | 1 |
| | D   1 to 0 | 1 | 1 | 0 | 0 | 1 | 1 | 1 | 1 | 1 |
| Boolean | B   0 to 1 | 0 | 1 | 1 | 0 | 1 | 1 | 1 | 1 | 1 |
| B1245678/ | S   1 to 1 | 1 | 1 | 0 | 0 | 1 | 1 | 1 | 1 | 1 |
| S0145678 | U   0 to 0 | 1 | 0 | 0 | 1 | 0 | 0 | 0 | 0 | 0 |
| even $t$ | D   1 to 0 | 0 | 0 | 1 | 1 | 0 | 0 | 0 | 0 | 0 |
| Boolean | B   0 to 1 | 0 | 0 | 0 | 0 | 0 | 1 | 1 | 0 | 0 |
| B56/S58 | S   1 to 1 | 0 | 0 | 0 | 0 | 0 | 1 | 0 | 0 | 1 |
| odd $t$ | U   0 to 0 | 1 | 1 | 1 | 1 | 1 | 0 | 0 | 1 | 1 |
| | D   1 to 0 | 1 | 1 | 1 | 1 | 1 | 0 | 1 | 1 | 0 |
| Real | B   0 to 1 | 0 | 0.1030 | 0.1332 | 0 | 0.0892 | 0.0489 | 0.0182 | 0.0033 | 0.0004 |
| B1245678/ | S   1 to 1 | 0.0090 | 0.0371 | 0 | 0 | 0.0628 | 0.0311 | 0.0125 | 0.0040 | 0.0147 |
| S0145678 | U   0 to 0 | 0.1562 | 0 | 0 | 0.1277 | 0 | 0 | 0 | 0 | 0 |
| even $t$ | D   1 to 0 | 0 | 0 | 0.0703 | 0.0785 | 0 | 0 | 0 | 0 | 0 |
| Real | B   0 to 1 | 0 | 0 | 0 | 0 | 0 | 0.0756 | 0.0622 | 0 | 0 |
| B56/S58 | S   1 to 1 | 0 | 0 | 0 | 0 | 0 | 0.1309 | 0 | 0 | 0.0408 |
| odd $t$ | U   0 to 0 | 0.1049 | 0.0189 | 0.0416 | 0.0558 | 0.0652 | 0 | 0 | 0.0303 | 0.0062 |
| | D   1 to 0 | 0.0001 | 0.0033 | 0.0250 | 0.0690 | 0.0999 | 0 | 0.1045 | 0.0656 | 0 |

Table 4. This table shows the twenty nearest neighbours of the Game of Life, the target rule B3/S23. Duplicate rules occur when a rule must be changed to avoid infinity.

| Rank | Rule | Anti-Infinity Rule Change | Duplicate Rules | Distance from B3/S23 | |
|---|---|---|---|---|---|
| | | | | Real | Boolean |
| 1 | B3/S23 — Target Rule | | 1 & 2 | 0.0000 | 0.0000 |
| 2 | B0123478/S01234678 | B3/S23 | 1 & 2 | 0.0146 | 0.0000 |
| 3 | B0123478/S1234678 | B38/S23 | 3 & 5 | 0.0269 | 1.4142 |
| 4 | B3/S238 | | | 0.0298 | 1.4142 |
| 5 | B38/S23 | | 3 & 5 | 0.0313 | 1.4142 |
| 6 | B38/S238 | | | 0.0413 | 2.0000 |
| 7 | B0123478/S0234678 | B37/S23 | 7 & 9 | 0.0553 | 1.4142 |
| 8 | B378/S23 | | 8 & 10 | 0.0625 | 2.0000 |
| 9 | B37/S23 | | 7 & 9 | 0.0664 | 1.4142 |
| 10 | B0123478/S234678 | B378/S23 | 8 & 10 | 0.0678 | 2.0000 |
| 11 | B37/S238 | | | 0.0838 | 2.0000 |
| 12 | B378/S238 | | | 0.0890 | 2.4495 |
| 13 | B3/S237 | | 13 & 15 | 0.0925 | 1.4142 |
| 14 | B3/S2378 | | | 0.0937 | 2.0000 |
| 15 | B023478/S01234678 | B3/S237 | 13 & 15 | 0.0949 | 1.4142 |
| 16 | B48/S23678 | | | 0.0963 | 3.4641 |
| 17 | B468/S236 | | | 0.0970 | 3.1623 |
| 18 | B03478/S01235678 | B4/S2367 | 18 & 34 | 0.0972 | 2.8284 |
| 19 | B478/S238 | | | 0.0976 | 3.1623 |
| 20 | B01468/S034678 | B367/S1356 | 20 & 136 | 0.0979 | 3.4641 |

Table 5. This table shows the twenty nearest neighbours of the target rule B03/S23 in the 72-dimensional probability space.

|      |                      | Anti-Strobing Rule Change |              | Distance from B03/S23 |         |
| ---: | -------------------- | ------------------------- | ------------ | ---: | ---: |
| Rank | Rule                 | Even Rule                 | Odd Rule     | Real | Boolean |
| 1    | B03/S23 — Target Rule | B1245678/S0145678        | B56/S58      | 0.0000 | 0.0000 |
| 2    | B038/S23             | B124567/S0145678          | B56/S058     | 0.0215 | 1.4142 |
| 3    | B037/S23             | B124568/S0145678          | B56/S158     | 0.0237 | 1.4142 |
| 4    | B0378/S23            | B12456/S0145678           | B56/S0158    | 0.0317 | 2.0000 |
| 5    | B037/S023            | B124568/S145678           | B568/S158    | 0.0365 | 2.0000 |
| 6    | B0378/S023           | B12456/S145678            | B568/S0158   | 0.0452 | 2.4495 |
| 7    | B03/S023             | B1245678/S145678          | B568/S58     | 0.0468 | 1.4142 |
| 8    | B038/S023            | B124567/S145678           | B568/S058    | 0.0525 | 2.0000 |
| 9    | B036/S23             | B124578/S0145678          | B56/S258     | 0.0795 | 1.4142 |
| 10   | B0368/S23            | B12457/S0145678           | B56/S0258    | 0.0816 | 2.0000 |
| 11   | B03678/S23           | B1245/S0145678            | B56/S01258   | 0.0889 | 2.4495 |
| 12   | B0367/S23            | B12458/S0145678           | B56/S1258    | 0.0890 | 2.0000 |
| 13   | B038/S236            | B124567/S014578           | B256/S058    | 0.1020 | 2.0000 |
| 14   | B036/S023            | B124578/S145678           | B568/S258    | 0.1025 | 2.0000 |
| 15   | B03/S236             | B1245678/S014578          | B256/S58     | 0.1042 | 1.4142 |
| 16   | B0368/S023           | B12457/S145678            | B568/S0258   | 0.1045 | 2.4495 |
| 17   | B0378/S236           | B12456/S014578            | B256/S0158   | 0.1094 | 2.4495 |
| 18   | B03/S0236            | B1245678/S14578           | B2568/S58    | 0.1103 | 2.0000 |
| 19   | B03678/S023          | B1245/S145678             | B568/S01258  | 0.1120 | 2.8284 |
| 20   | B037/S236            | B124568/S014578           | B256/S158    | 0.1125 | 2.0000 |